\documentclass[review,12pt]{elsarticle}

\usepackage[utf8]{inputenc}
\usepackage[T1]{fontenc}
\usepackage{amsmath,amssymb,amsfonts}
\usepackage{graphicx}
\usepackage{float}        
\usepackage[section]{placeins} 

\usepackage{booktabs}
\usepackage{array}
\usepackage{pifont}
\usepackage{xcolor}
\usepackage[hidelinks]{hyperref}

\biboptions{numbers,sort&compress}

\newcommand{\cmark}{\ding{51}}
\newcommand{\xmark}{\ding{55}}

\journal{Pattern Recognition}

\begin{document}

\begin{frontmatter}

\title{GMBFormer: An NDVI-Guided Global Memory Bank Transformer for Urban Green-Space Extraction from Ultra-High-Resolution Imagery}

\author[inst1]{Hao Lei}
\author[inst1]{Xi Cheng\corref{cor1}\fnref{fn1}}
\ead{chengxi13@cdut.edu.cn}
\author[inst1]{Chenlu Shu}
\author[inst1]{Zhiheng Chen}
\author[inst1]{Zhengjie Duan}
\author[inst2]{Haoyu Wang}
\author[inst2]{Zhanfeng Shen}
\address[inst1]{College of Geophysics, Chengdu University of Technology, Chengdu 610059, China}
\address[inst2]{National Engineering Research Center for Geomatics, Aerospace Information Research Institute, Chinese Academy of Sciences, and University of Chinese Academy of Sciences, Beijing 100101, China}
\cortext[cor1]{Corresponding author.}
\fntext[fn1]{ORCID: \href{https://orcid.org/0000-0002-0587-8651}{0000-0002-0587-8651}.}


\begin{abstract}
Urban green-space extraction from ultra-high-resolution (UHR) imagery is commonly performed patch by patch, which limits semantic reuse among spatially separated but visually similar vegetation patterns. Directly injecting the Normalized Difference Vegetation Index (NDVI) into red--green--blue (RGB) backbones can also blur the roles of visual appearance learning and physical vegetation confidence. We propose GMBFormer, a SegFormer-based framework that replaces adjacency-driven feature propagation with selective, similarity-driven prototype retrieval. Only RGB channels enter the backbone and decoder, while NDVI is decoupled as a physics-informed gate that admits high-confidence vegetation descriptors into a compact global memory bank through momentum updates. During training and inference, the current patch queries stored prototypes through memory-mediated cross-attention, and the retrieved response is integrated with bounded overhead. Experiments use a self-constructed Chengdu UHR dataset with 7,700 labeled $512 \times 512$ patches and two reduced-label settings derived from the public International Society for Photogrammetry and Remote Sensing (ISPRS) Potsdam dataset. Under the same training and evaluation protocol, GMBFormer obtains mean intersection over union (mIoU)/mean Dice (mDice) scores of 89.25\%/94.31\%, 92.17\%/95.92\%, and 83.72\%/90.86\%, respectively, improving the controlled SegFormer-B4 baseline in each setting. Ablation studies indicate that decoupled NDVI admission, memory retrieval, capacity, and momentum jointly shape the final performance.
\end{abstract}


\end{frontmatter}

\section{Introduction}
\label{sec:introduction}

Urban green spaces support ecological assessment, planning, and public health by regulating temperature and stormwater and sustaining biodiversity \citep{Bertram2015,Kondo2018,Huang2025}. Ultra-high-resolution (UHR) remote sensing imagery enables fine-grained mapping of these spaces, but its spatial detail also exposes segmentation models to fragmented vegetation boundaries, shadows, artificial green-like surfaces, and strong intra-class variability \citep{Derkzen2015}.

Deep segmentation models, from convolutional neural networks (CNNs) \citep{Long2015,Ronneberger2015,Chen2018} to Transformers and recent state space models \citep{Dosovitskiy2021,Liu2021,GuDao2024,Liu2024VMamba,Chen2024RSMamba}, have improved remote sensing interpretation. SegFormer \citep{Xie2021}, in particular, offers a strong balance between hierarchical representation and efficiency. Yet UHR mapping is still commonly trained and inferred on cropped patches: once a patch is processed, its feature evidence is discarded, and similar green-space patterns elsewhere can only be recalled implicitly through shared weights. This is limiting because urban greenery may appear as courtyards, roadside strips, compact tree crowns, or large parks across disconnected crops. Enlarging the receptive field within one crop cannot recover vegetation evidence that is absent from that crop but semantically repeated elsewhere.

This patch-wise formulation creates three difficulties. First, spatial continuity is broken across crops, and sliding-window propagation may transfer misleading context because adjacency does not imply semantic similarity. Second, urban vegetation varies in illumination, canopy structure, phenology, and scale, while shadows and green-like impervious surfaces often cause confusion \citep{Huang2025}. Third, the Normalized Difference Vegetation Index (NDVI) provides physical vegetation evidence, but direct concatenation, multimodal fusion, or feature-level interaction with red--green--blue (RGB) imagery \citep{Sa2018,Diakogiannis2020,Zhou2022a,Huang2025} can entangle RGB appearance learning with vegetation-confidence estimation.

To address these issues, we propose GMBFormer (Global Memory Bank-enhanced Transformer), which externalizes part of the learned vegetation knowledge into a compact, queryable memory. During memory writing, NDVI is detached from backbone optimization and used only as a physics-informed admission gate for high-confidence green-space prototypes. During memory reading, current RGB-derived features query the stored prototypes through cross-attention, so enhancement is driven by semantic similarity rather than patch adjacency. The memory has fixed capacity and is frozen at inference time, keeping retrieval explicit and bounded.

The main contributions of this work are summarized as follows:
\begin{enumerate}
  \item We introduce an NDVI-guided global memory bank for cross-patch semantic reuse, using NDVI only for quality-controlled memory admission while keeping RGB representation learning and decoding intact.
  \item We design a memory-mediated cross-attention module that lets each patch retrieve vegetation prototypes by semantic similarity, improving non-contiguous green-space recognition without explicit inter-patch communication.
  \item We validate GMBFormer on a self-constructed Chengdu UHR dataset and two reduced-label ISPRS Potsdam settings; ablations analyze admission gating, retrieval, memory capacity, and EMA momentum.
\end{enumerate}

\section{Related Work}
\label{sec:related_work}

\subsection{Urban Green Space Extraction from Remote Sensing Imagery}

Urban green space extraction supports ecological assessment and sustainable planning \citep{Bertram2015,Kondo2018}. Traditional methods relied on pixel-level classification, object-based image analysis, and handcrafted spectral indices, with NDVI serving as a long-standing vegetation discriminator \citep{Myint2011,Puissant2014,Rouse1974,Tucker1979}. Deep models have improved accuracy, but UHR urban scenes remain difficult because vegetation varies across species, illumination, and canopy structure, while shadows and dark impervious surfaces can mimic green-space appearance \citep{Derkzen2015,Huang2025}. These characteristics make the task more specific than generic encoder--decoder segmentation.

\subsection{Transformer, State Space Model, and Efficient Segmentation}

Transformer-based architectures such as Vision Transformer (ViT), Swin Transformer, and SegFormer have strengthened dense prediction through long-range modeling and hierarchical representation \citep{Dosovitskiy2021,Liu2021,Xie2021}, and remote-sensing variants such as UNetFormer and DC-Swin confirm their value for high-resolution land-cover mapping \citep{Wang2022a,Wang2022b}. State space model (SSM)-based models, including Mamba, VMamba, and RSMamba, further improve sequence modeling efficiency \citep{GuDao2024,Liu2024VMamba,Chen2024RSMamba}, while auxiliary geometric priors have also been explored for aerial segmentation \citep{Peng2023MSINet}. However, UHR imagery is still commonly processed as cropped patches because of memory constraints \citep{Ding2021LANet}. Architectures such as GLNet, FCtL, ISDNet, and WiCoNet introduce global branches, contextual retrieval, shallow-detail distillation, or neighbor-window attention \citep{Chen2019,Li2021a,Guo2022ISDNet,Ding2022}. These strategies improve local or adjacent-patch consistency, but their context is usually tied to the current image, a downsampled scene representation, or nearby windows rather than a persistent memory queried by non-contiguous patches.

\subsection{Vegetation Index Utilization in Remote Sensing Segmentation}

NDVI provides a physically grounded cue for separating vegetation from spectrally similar non-vegetated surfaces \citep{Rouse1974,Tucker1979,Huang2025}. Existing deep models usually inject vegetation indices through channel concatenation \citep{Sa2018,Diakogiannis2020}, multimodal fusion \citep{Tong2020,Zhou2022a}, or feature-level interaction \citep{Huang2025}. These strategies can improve complementarity, but they often treat NDVI as another learnable feature rather than separating RGB appearance encoding from vegetation-confidence estimation. Related red--green--blue-depth (RGB-D) segmentation work also shows that modalities with different physical meanings require careful interaction rather than naive mixing \citep{Zhou2022a}. In our setting, such entanglement may weaken pretrained RGB priors and obscure NDVI as a physical confidence signal.

\subsection{Prototype and Memory-Based Segmentation Methods}

Prototype and memory methods offer reusable representations beyond standard feed-forward extraction. PANet, ASGNet, ProtoSeg, and Attentional Prototype Inference (API) use prototypes to improve class-level discrimination or few-shot segmentation \citep{Wang2019,Li2021b,Zhou2022b,Sun2023API}, but their prototypes are typically image-local or task-local. Video object segmentation methods such as STM, STCN, and XMem store historical frame features for attention-based retrieval \citep{Oh2019,Cheng2021,Cheng2022}, and static segmentation methods also use learnable or retrieval-augmented memory for feature enhancement and domain generalization \citep{Guo2023ExternalAttention,Kim2022}. Yet these methods do not address how a remote-sensing model should decide which RGB--NDVI patches are reliable enough to enter a persistent memory. This admission problem is critical in urban green-space extraction because ambiguous shadows, mixed pixels, or artificial surfaces can pollute memory if all features are stored solely by learned similarity.

In summary, three gaps remain: UHR segmentation lacks a globally shared semantic memory for non-contiguous cross-patch retrieval; RGB--NDVI fusion often entangles appearance and physical-index roles; and existing memory methods rarely use domain priors to control memory quality. GMBFormer addresses these gaps by coupling NDVI-guided admission with similarity-based prototype retrieval.

\section{Methodology}
\label{sec:methodology}

\subsection{Overall Framework}
\label{subsec:overall_framework}

\begin{figure}[H]
  \centering
  \includegraphics[width=1\linewidth]{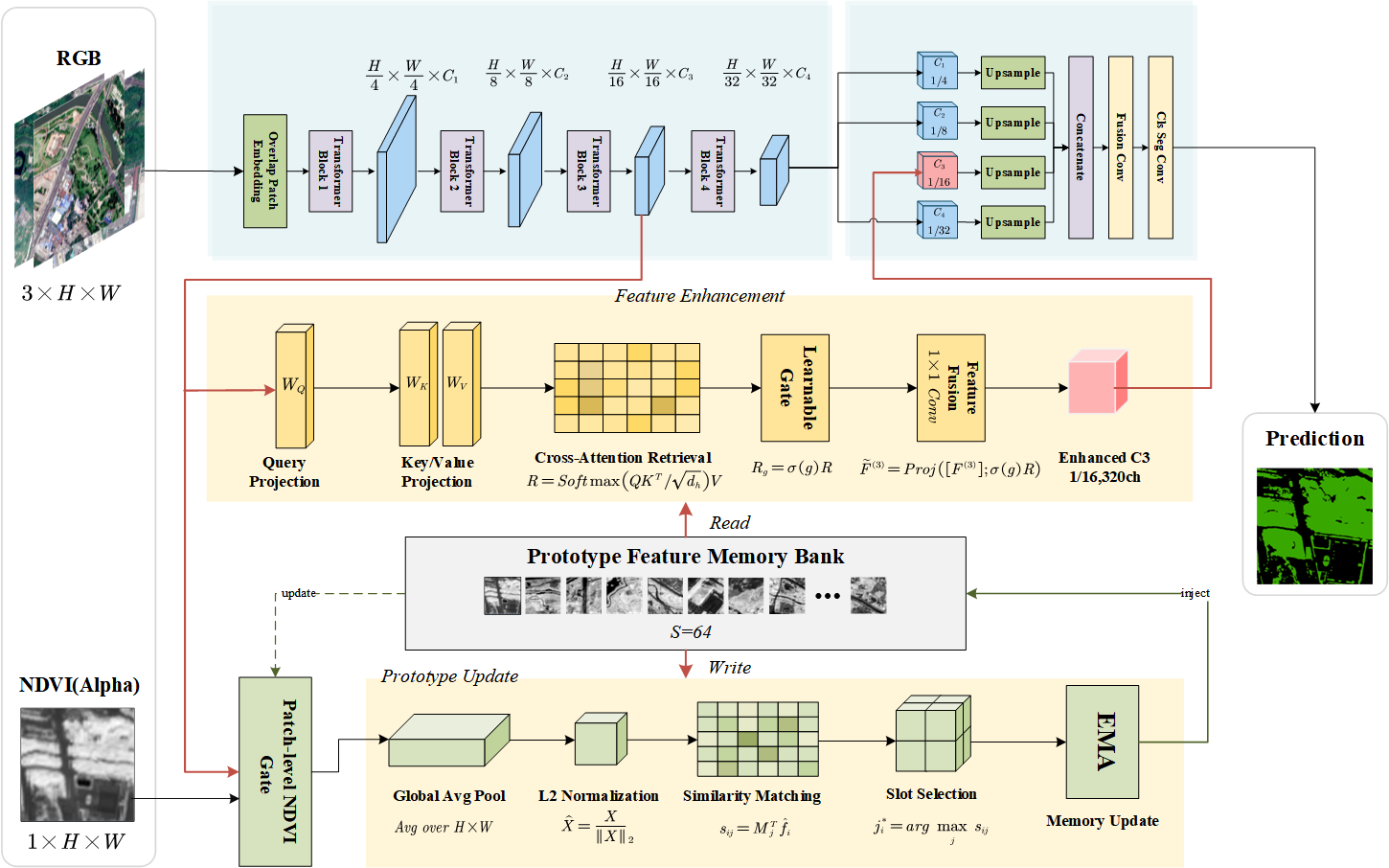}
  \caption{Overall architecture of GMBFormer. Only RGB is fed into the MiT-B4 backbone and decoder, while NDVI in the alpha channel gates training-time memory updates. The memory bank stores $S=64$ Stage--3 vegetation prototypes, retrieves them through cross-attention, and fuses the response before SegFormer-style decoding.}
  \label{fig:overall_architecture}
\end{figure}

Fig.~\ref{fig:overall_architecture} shows the architecture of GMBFormer. The four-channel RGB--NDVI input is decomposed into an RGB stream and an NDVI stream: RGB channels are processed by the MiT-B4 backbone, whereas NDVI is detached from gradient optimization and used only to decide which training patches can write into the Global Memory Bank (GMB). Thus, NDVI controls memory quality without becoming a learnable image feature in the backbone or decoder. The GMB writes high-confidence vegetation prototypes by exponential moving average (EMA) during training, and reads stored prototypes through cross-attention during both training and inference. The retrieved response is fused into Stage--3 features through a learnable gate, with $S=64$ prototype slots by default.

For an input $\mathbf{I}=[\mathbf{I}_{\text{RGB}};\mathbf{N}] \in \mathbb{R}^{4 \times H \times W}$, only $\mathbf{I}_{\text{RGB}}$ is forwarded to the backbone:
\begin{equation}
\{\mathbf{F}^{(1)},\mathbf{F}^{(2)},\mathbf{F}^{(3)},\mathbf{F}^{(4)}\}
= \text{MiT-B4}(\mathbf{I}_{\text{RGB}}),
\label{eq:backbone_features}
\end{equation}
where $C_1{=}64$, $C_2{=}128$, $C_3{=}320$, and $C_4{=}512$. The NDVI channel is used only for the memory-admission decision in Eq.~\eqref{eq:admission_gate}; it is not concatenated with RGB features, passed to the decoder, or optimized through the segmentation loss. The GMB enhances $\mathbf{F}^{(3)} \in \mathbb{R}^{C_3 \times H/16 \times W/16}$, which balances spatial detail and semantic abstraction for prototype matching, and the enhanced feature $\tilde{\mathbf{F}}^{(3)}$ replaces $\mathbf{F}^{(3)}$ in the SegFormer decoder.

\subsection{NDVI-Guided Memory Write}

The global memory bank $\mathbf{M} \in \mathbb{R}^{S \times C_3}$ maintains $S$ $\ell_2$-normalized prototype vectors of dimension $C_3$ for cosine matching. It is initialized with normalized random Gaussian vectors and updated only through the momentum mechanism below, without gradient-based optimization.

Not every patch should contribute to the memory. Patches dominated by non-vegetated land cover can contaminate green-space prototypes, so the mean NDVI value of each training patch $i$ is used as an admission gate:
\begin{equation}
\bar{n}_i = \frac{1}{H_n W_n} \sum_{u,v} \mathbf{N}_i(u,v).
\label{eq:mean_ndvi}
\end{equation}
where $H_n$ and $W_n$ denote the spatial dimensions of the NDVI gate map. Thus, NDVI affects whether an RGB-derived prototype is written to memory, but it does not provide spatial features for pixel-level prediction.

A patch is admitted to update the memory bank only if its mean NDVI exceeds a predefined threshold $\tau_{\text{norm}}$:
\begin{equation}
\text{admit}(i) = \mathbb{1}[\bar{n}_i > \tau_{\text{norm}}].
\label{eq:admission_gate}
\end{equation}

For each admitted patch, Stage--3 features are pooled by global average pooling (GAP) and normalized:
\begin{equation}
\mathbf{f}_i = \text{GAP}(\mathbf{F}_i^{(3)}) \in \mathbb{R}^{C_3}, \quad
\hat{\mathbf{f}}_i = \frac{\mathbf{f}_i}{\|\mathbf{f}_i\|_2}.
\label{eq:patch_descriptor}
\end{equation}
The most similar slot is selected by cosine similarity,
\begin{equation}
j^* = \arg\max_{j \in \{1,\ldots,S\}} \; \mathbf{M}_j^\top \hat{\mathbf{f}}_i.
\label{eq:prototype_matching}
\end{equation}
and updated by EMA with momentum $\alpha$:
\begin{equation}
\mathbf{M}_{j^*} \leftarrow
\frac{\alpha \, \mathbf{M}_{j^*} + (1-\alpha) \, \hat{\mathbf{f}}_i}
{\|\alpha \, \mathbf{M}_{j^*} + (1-\alpha) \, \hat{\mathbf{f}}_i\|_2}.
\label{eq:ema_update}
\end{equation}
Updates are performed without gradient computation, and the final normalization keeps prototypes on the unit hypersphere.

\subsection{Memory Read and Gated Fusion}

Let $H_3=H/16$ and $W_3=W/16$. Queries are projected from $\mathbf{F}^{(3)}$, whereas keys and values are projected from the memory bank:
\begin{equation}
\mathbf{Q} = \mathbf{W}_Q \ast \mathbf{F}^{(3)}
\in \mathbb{R}^{C_3 \times H_3 \times W_3}, \quad
\mathbf{K} = \mathbf{W}_K \, \mathbf{M}^\top \in \mathbb{R}^{C_3 \times S}, \quad
\mathbf{V} = \mathbf{W}_V \, \mathbf{M}^\top \in \mathbb{R}^{C_3 \times S}.
\label{eq:qkv_projection}
\end{equation}
For each head $h$ and spatial position $p$, memory attention and retrieval are computed as
\begin{equation}
\begin{aligned}
a_{p,m}^{(h)} =
\frac{\exp\!\left(\mathbf{Q}_{p}^{(h)} \cdot \mathbf{K}_{m}^{(h)\top}/\sqrt{d_h}\right)}
{\sum_{m'=1}^{S} \exp\!\left(\mathbf{Q}_{p}^{(h)} \cdot \mathbf{K}_{m'}^{(h)\top}/\sqrt{d_h}\right)}, \quad
\mathbf{R}_{p}^{(h)} = \sum_{m=1}^{S} a_{p,m}^{(h)} \, \mathbf{V}_{m}^{(h)} .
\end{aligned}
\label{eq:memory_attention}
\end{equation}
The head outputs are concatenated and reshaped to $\mathbf{R} \in \mathbb{R}^{C_3 \times H_3 \times W_3}$. A scalar gate $g$ initialized to $-3$ modulates the memory response:
\begin{equation}
\tilde{\mathbf{F}}^{(3)} =
\text{Proj}\!\left(\left[\mathbf{F}^{(3)} \;;\; \sigma(g) \cdot \mathbf{R}\right]\right),
\label{eq:gated_fusion}
\end{equation}
where $\text{Proj}(\cdot)$ is a $1 \times 1$ convolutional projection from $2C_3$ to $C_3$. Since $\sigma(-3)\approx0.05$, the memory pathway starts weak and learns how much retrieved information should enter the decoder.

\subsection{Decoder, Objective, and Complexity}

The decoder follows SegFormer \citep{Xie2021}: $\{\mathbf{F}^{(1)},\mathbf{F}^{(2)},\tilde{\mathbf{F}}^{(3)},\mathbf{F}^{(4)}\}$ are projected by multilayer perceptron (MLP) layers, upsampled to $H/4 \times W/4$, concatenated, fused, and classified into $K$ channels. We use $K{=}2$ for binary green-space extraction and $K{=}3$ for the Potsdam background/low-vegetation/tree setting, with bilinear upsampling to the original resolution during inference.

The model is optimized with cross-entropy (CE) and Dice losses:
\begin{equation}
\mathcal{L} = \mathcal{L}_{\text{CE}} + \lambda \, \mathcal{L}_{\text{Dice}},
\label{eq:total_loss}
\end{equation}
with $\lambda=0.5$. Memory writing in Eqs.~\eqref{eq:mean_ndvi}--\eqref{eq:ema_update} is active only during training; at inference time, the bank is frozen.

The memory bank itself is small: with $S{=}64$ and $C_3{=}320$, it occupies about 0.08~MB in float32. The read operation scales as
\begin{equation}
\mathcal{O}(H_3 W_3 C_3^2 + 2S C_3^2 + 2H_3 W_3 S C_3 + 2H_3 W_3 C_3^2).
\label{eq:complexity}
\end{equation}
For $512 \times 512$ inputs, $H_3{=}W_3{=}32$, $S{=}64$, and $C_3{=}320$; the attention terms require about $4.2\times10^7$ multiply--accumulate operations, and the Q/K/V projections and fusion add approximately $5C_3^2 \approx 0.51$~M parameters.

\section{Experiments}
\label{sec:experiments}

\subsection{Datasets and Study Areas}
\label{subsec:datasets}

\begin{figure}[H]
  \centering
  \includegraphics[width=1\linewidth]{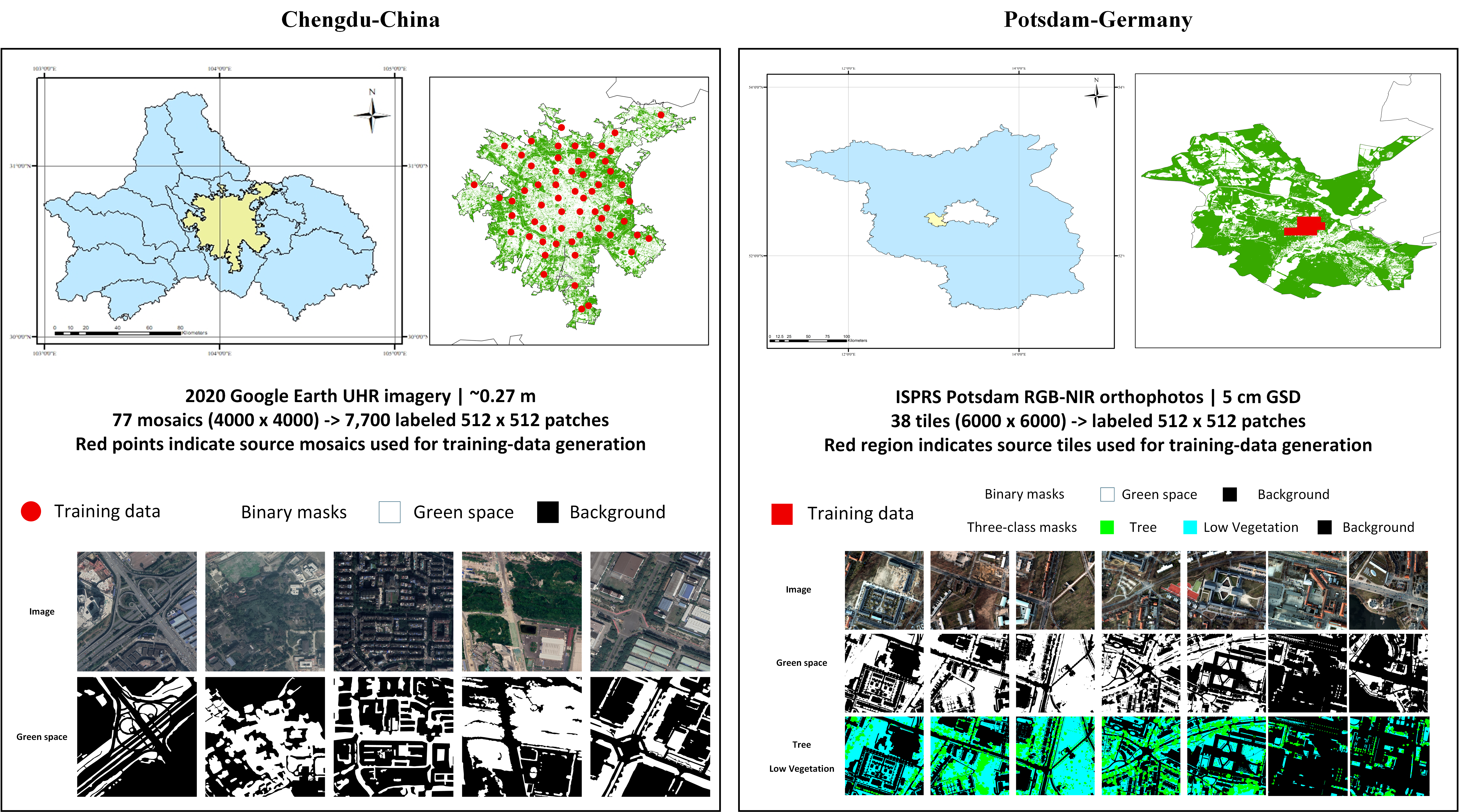}
  \caption{Study areas and representative samples for the Chengdu and ISPRS Potsdam datasets, including Chengdu source mosaics and binary labels after split-before-crop generation, and Potsdam source tiles with binary and three-class label conversions.}
  \label{fig:study_area_dataset_overview}
\end{figure}

Fig.~\ref{fig:study_area_dataset_overview} summarizes the two study areas and sample construction. It shows the Chengdu study area, mapped green-space distribution, source Google Earth mosaics, and representative binary training samples, together with the Potsdam dataset region, source tile region, and binary/three-class label conversions.

Chengdu UHR Dataset. The Chengdu dataset was built from 77 Google Earth RGB mosaics acquired in 2020 at approximately 0.27~m spatial resolution. Each source mosaic was prepared as a $4000 \times 4000$ image after clipping, stitching, color correction, reprojection to the World Geodetic System 1984 (WGS-84), and temporal consistency checking with Sentinel-2 observations. Manual annotation used high-resolution imagery and reference data; trees, lawns, and natural green spaces were merged into foreground, non-green-space areas into background, and ambiguous pixels into ignore (255). To avoid overlap leakage, source mosaics or spatial blocks were first assigned to training or validation, and 25\% overlapping $512 \times 512$ crops were generated only within the assigned split. This split-before-crop protocol produced 7,700 labeled patches, including 6,160 training and 1,540 validation samples.

NDVI for Chengdu was derived from Sentinel-2 imagery rather than Google Earth RGB images. Twelve low-cloud Sentinel-2 observations in 2020 were selected for each of nine large sub-regions through Google Earth Engine, and NDVI was computed from the near-infrared (NIR) and red bands. The 10~m NDVI layers were aligned with the UHR grid, bilinearly upsampled, mapped from $[-1,1]$ to $[0,255]$, and stored in the alpha channel. For GMBFormer, this alpha channel is only a memory-admission gate and never enters RGB feature learning, decoder fusion, or the segmentation loss. This keeps RGB appearance and NDVI confidence distinct, although the 10~m-to-UHR mismatch can introduce mixed-pixel uncertainty.

ISPRS Potsdam Dataset. We further evaluate on the ISPRS 2D Semantic Labeling Potsdam dataset, which contains 38 aerial true orthophoto tiles at 5~cm ground sampling distance with RGB and NIR channels \citep{Rottensteiner2012,Rottensteiner2014}. Each tile is $6000 \times 6000$ pixels; following the standard split, 24 tiles are used for training and 14 for testing. We construct two reduced-label settings for green-space extraction: a binary setting that merges \textit{low vegetation} and \textit{tree} as green space, and a three-class setting that merges non-vegetation categories into \textit{background} while retaining \textit{low vegetation} and \textit{tree}. These task-specific conversions are intended for within-paper comparisons among models retrained under the same label mappings, not direct comparison with published six-class Potsdam results. NDVI is computed from NIR and red bands, stored as the alpha channel, and cropped with the RGB tiles into $512 \times 512$ patches.

\subsection{Implementation Details}

All experiments were implemented in MMSegmentation 1.x \citep{OpenMMLab2023} with PyTorch 2.x and trained on a single NVIDIA RTX 3060 graphics processing unit (GPU) with 12~GB memory. GMBFormer and SegFormer-B4 use MiT-B4 with a SegFormer-style decoder; Swin-UPerNet uses Swin-Base, Mask2Former uses ResNet-50 with a Mask2Former head, and DeepLabV3 uses ResNet-50-D8 with atrous spatial pyramid pooling (ASPP). ImageNet-pretrained weights are used where available, while GMB modules are initialized from scratch.

All models use the same data split, $512 \times 512$ crop size, augmentation pipeline, 320,000 training iterations, 32,000-iteration validation interval, and evaluation protocol. Augmentation includes random resizing (0.5--2.0), random cropping with at most 75\% single-class ratio, horizontal/vertical flipping, and photometric distortion. Backbone choices follow each method's standard configuration. The controlled architectural comparison is GMBFormer versus SegFormer-B4, which share the MiT-B4 backbone and SegFormer-style decoder; the other methods serve as representative references from mask-classification, hierarchical-Transformer, and convolutional segmentation families. All baseline results reported in this study are produced by retraining the models under the same protocol and, for Potsdam, the same reduced-label mappings. GMBFormer, SegFormer-B4, and Swin-UPerNet use AdamW \citep{Loshchilov2019}; Mask2Former follows its standard AdamW setting, and DeepLabV3 uses stochastic gradient descent (SGD). Unless otherwise specified, GMBFormer uses $S=64$, $\alpha=0.99$, $\tau_{\text{raw}}=0.2$, and $n_h=8$. NDVI thresholds are reported on the raw $[-1,1]$ scale and mapped to the stored $[0,1]$ scale by $\tau_{\text{norm}}=(\tau_{\text{raw}}+1)/2$; thus, $\tau_{\text{raw}}=0.2$ corresponds to $\tau_{\text{norm}}=0.6$. The checkpoint with the best validation mIoU is used for reporting.

\subsection{Evaluation Metrics}

We report overall accuracy (aAcc), mean intersection over union (mIoU), mean accuracy (mAcc), mean Dice (mDice), mean precision (mPrecision), and mean recall (mRecall), with mIoU as the primary metric. Dice and the F1-score are equivalent when $\beta=1$, so F-score is not reported as a separate column. Ignore-labeled pixels (255) are excluded from all computations. For binary settings, foreground IoU and Dice refer to the green-space class unless otherwise specified. For the Potsdam binary setting, class-wise IoU is reported for background and green space; for the three-class setting, IoU is reported for background, low vegetation, and tree to verify whether gains extend to both vegetation subclasses.

\subsection{Comparison with Representative Segmentation Methods}

We report results for Mask2Former, Swin-UPerNet, DeepLabV3, and SegFormer-B4 to cover mask classification, hierarchical Transformer decoding, dilated convolution, and the RGB-only backbone family of our method. Because these methods use different standard backbones, performance differences among all methods should be interpreted as representative method comparisons; the controlled comparison for the proposed memory design is between GMBFormer and SegFormer-B4. For Potsdam, all methods are retrained under the custom binary or three-class label mappings described in Section~\ref{subsec:datasets}; no numbers are taken from published six-class leaderboards.

\subsubsection{Chengdu UHR Dataset}

Table~\ref{tab:chengdu_validation} reports the quantitative comparison on the Chengdu UHR validation set.

\begin{table}[H]
\centering
\caption{Quantitative comparison on the Chengdu UHR validation set.}
\label{tab:chengdu_validation}
\scriptsize
\setlength{\tabcolsep}{3pt}
\resizebox{\textwidth}{!}{%
\begin{tabular}{llcccccc}
\toprule
Method & Backbone & aAcc & mIoU & mAcc & mDice & mPrecision & mRecall \\
\midrule
Mask2Former & ResNet-50 & 90.98 & 86.12 & 90.74 & 90.77 & 90.81 & 90.74 \\
Swin-UPerNet & Swin-Base & 92.98 & 87.41 & 92.83 & 92.84 & 92.85 & 92.83 \\
DeepLabV3 & ResNet-50-D8 & 91.78 & 84.50 & 91.59 & 91.59 & 91.59 & 91.59 \\
SegFormer-B4 & MiT-B4 & 92.99 & 87.40 & 92.71 & 92.83 & 92.98 & 92.71 \\
GMBFormer (Ours) & MiT-B4 & 94.45 & 89.25 & 94.25 & 94.31 & 94.38 & 94.25 \\
\bottomrule
\end{tabular}
}
\end{table}

GMBFormer obtains the highest values on all Chengdu validation metrics, with mIoU increasing from 87.40\% for the controlled SegFormer-B4 baseline to 89.25\%. The simultaneous gains in precision and recall suggest that NDVI-gated memory retrieval improves RGB-based vegetation discrimination rather than simply expanding the predicted green-space foreground.

\begin{figure}[H]
  \centering
  \includegraphics[width=1\linewidth]{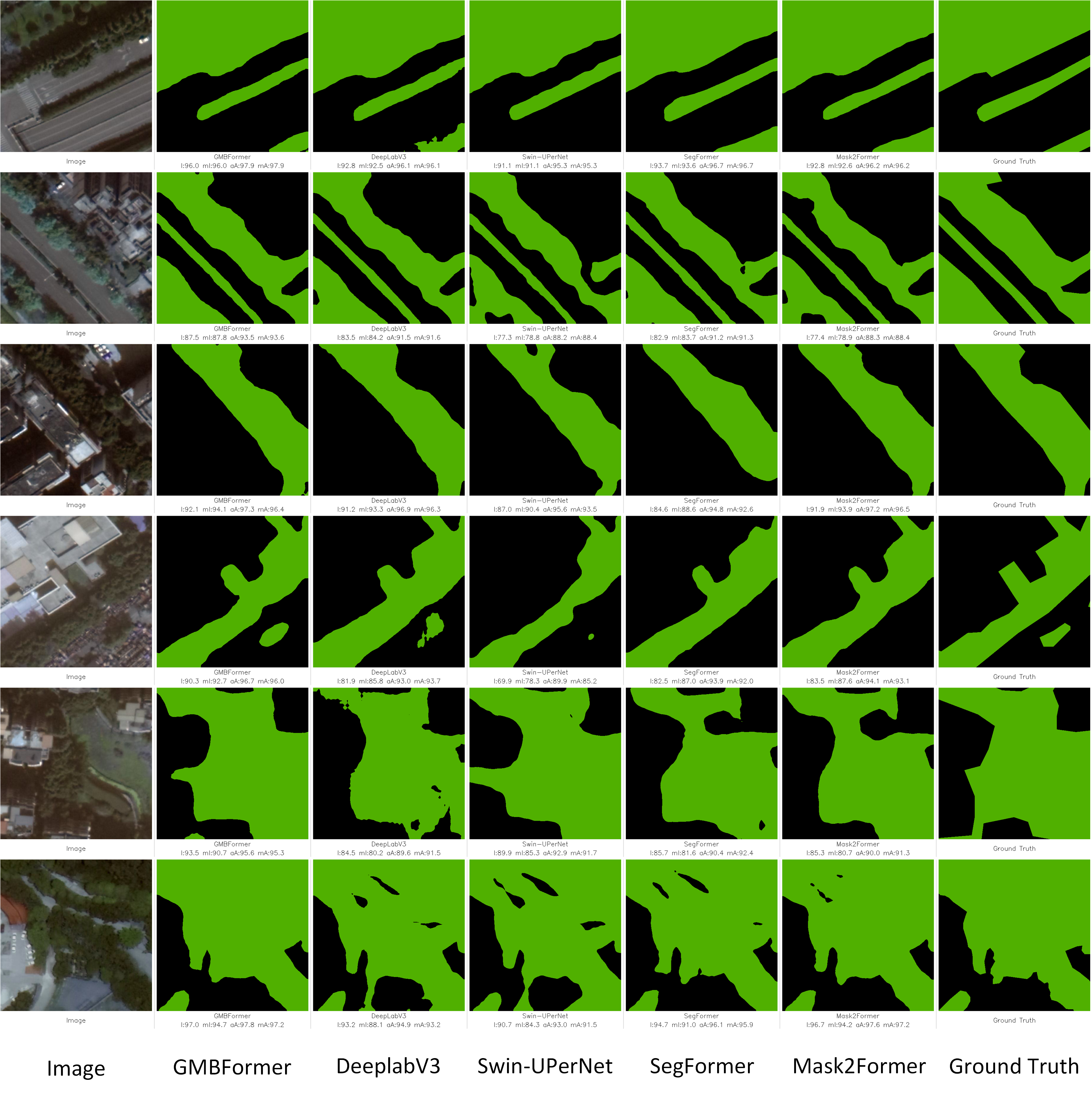}
  \caption{Qualitative comparison on Chengdu UHR binary green-space extraction results.}
  \label{fig:chengdu_qualitative_comparison}
\end{figure}

Fig.~\ref{fig:chengdu_qualitative_comparison} complements the Chengdu quantitative results by showing binary green-space extraction under dense built-up regions, fragmented vegetation, and shadow-affected areas. The visual comparison illustrates the practical role of retrieving reliable vegetation prototypes when a single patch contains only partial or ambiguous green-space evidence.

\subsubsection{ISPRS Potsdam Custom Binary Setting}

Table~\ref{tab:potsdam_binary} reports the custom binary Potsdam evaluation, where low vegetation and tree are merged into the green-space foreground and all non-vegetated categories are treated as background. This setting mirrors the Chengdu binary extraction task while introducing a public aerial dataset with different spatial resolution, sensor characteristics, and urban morphology.

\begin{table}[H]
\centering
\caption{Quantitative comparison on the custom ISPRS Potsdam binary setting. IoU$_{\text{bg}}$ and IoU$_{\text{green}}$ denote class-wise IoU for background and green space, respectively.}
\label{tab:potsdam_binary}
\scriptsize
\setlength{\tabcolsep}{3pt}
\resizebox{\textwidth}{!}{%
\begin{tabular}{llcccccccc}
\toprule
Method & Backbone & IoU$_{\text{bg}}$ & IoU$_{\text{green}}$ & aAcc & mIoU & mAcc & mDice & mPrecision & mRecall \\
\midrule
Mask2Former & ResNet-50 & 91.86 & 87.37 & 93.35 & 89.62 & 94.86 & 94.46 & 94.11 & 94.86 \\
Swin-UPerNet & Swin-Base & 93.70 & 90.09 & 95.99 & 91.89 & 95.73 & 95.76 & 95.80 & 95.73 \\
DeepLabV3 & ResNet-50-D8 & 93.12 & 89.35 & 95.64 & 91.23 & 95.52 & 95.40 & 95.29 & 95.52 \\
SegFormer-B4 & MiT-B4 & 93.58 & 89.95 & 95.92 & 91.76 & 95.70 & 95.70 & 95.69 & 95.70 \\
GMBFormer (Ours) & MiT-B4 & 93.89 & 90.45 & 96.13 & 92.17 & 95.95 & 95.92 & 95.88 & 95.95 \\
\bottomrule
\end{tabular}
}
\end{table}

Under this custom binary Potsdam setting, GMBFormer obtains the highest mIoU (92.17\%) and green-space IoU (90.45\%). The gains across the mean and class-wise metrics indicate that the decoupled gate-and-retrieval design remains useful on a public aerial benchmark with different spatial resolution, sensor characteristics, and urban morphology.

\subsubsection{ISPRS Potsdam Custom Three-Class Setting}

Table~\ref{tab:potsdam_three_class} reports the custom three-class Potsdam evaluation, which aggregates all non-vegetation categories as background and separates low vegetation and tree.

\begin{table}[H]
\centering
\caption{Quantitative comparison on the custom ISPRS Potsdam three-class setting. IoU$_{\text{bg}}$, IoU$_{\text{lv}}$, and IoU$_{\text{tree}}$ denote class-wise IoU for background, low vegetation, and tree, respectively.}
\label{tab:potsdam_three_class}
\scriptsize
\setlength{\tabcolsep}{3pt}
\resizebox{\textwidth}{!}{%
\begin{tabular}{llccccccccc}
\toprule
Method & Backbone & IoU$_{\text{bg}}$ & IoU$_{\text{lv}}$ & IoU$_{\text{tree}}$ & aAcc & mIoU & mAcc & mDice & mPrecision & mRecall \\
\midrule
Mask2Former & ResNet-50 & 92.90 & 76.58 & 75.20 & 92.51 & 81.56 & 89.42 & 89.63 & 89.85 & 89.42 \\
Swin-UPerNet & Swin-Base & 93.74 & 78.61 & 76.43 & 93.20 & 82.93 & 90.46 & 90.48 & 90.49 & 90.46 \\
DeepLabV3 & ResNet-50-D8 & 89.32 & 74.87 & 73.11 & 89.16 & 79.69 & 86.37 & 86.40 & 86.48 & 86.37 \\
SegFormer-B4 & MiT-B4 & 93.77 & 79.01 & 77.22 & 93.33 & 83.33 & 90.95 & 90.73 & 90.53 & 90.95 \\
GMBFormer (Ours) & MiT-B4 & 94.04 & 79.48 & 77.65 & 93.48 & 83.72 & 90.97 & 90.86 & 90.74 & 90.97 \\
\bottomrule
\end{tabular}
}
\end{table}

GMBFormer obtains the highest mIoU (83.72\%) and improves both vegetation subclasses, with IoU$_{\text{lv}}=79.48\%$ and IoU$_{\text{tree}}=77.65\%$. This pattern is consistent with the intended role of the memory bank: retrieved prototypes enhance vegetation structure without reducing the task to binary foreground--background separation.

\begin{figure}[H]
  \centering
  \includegraphics[width=1\linewidth]{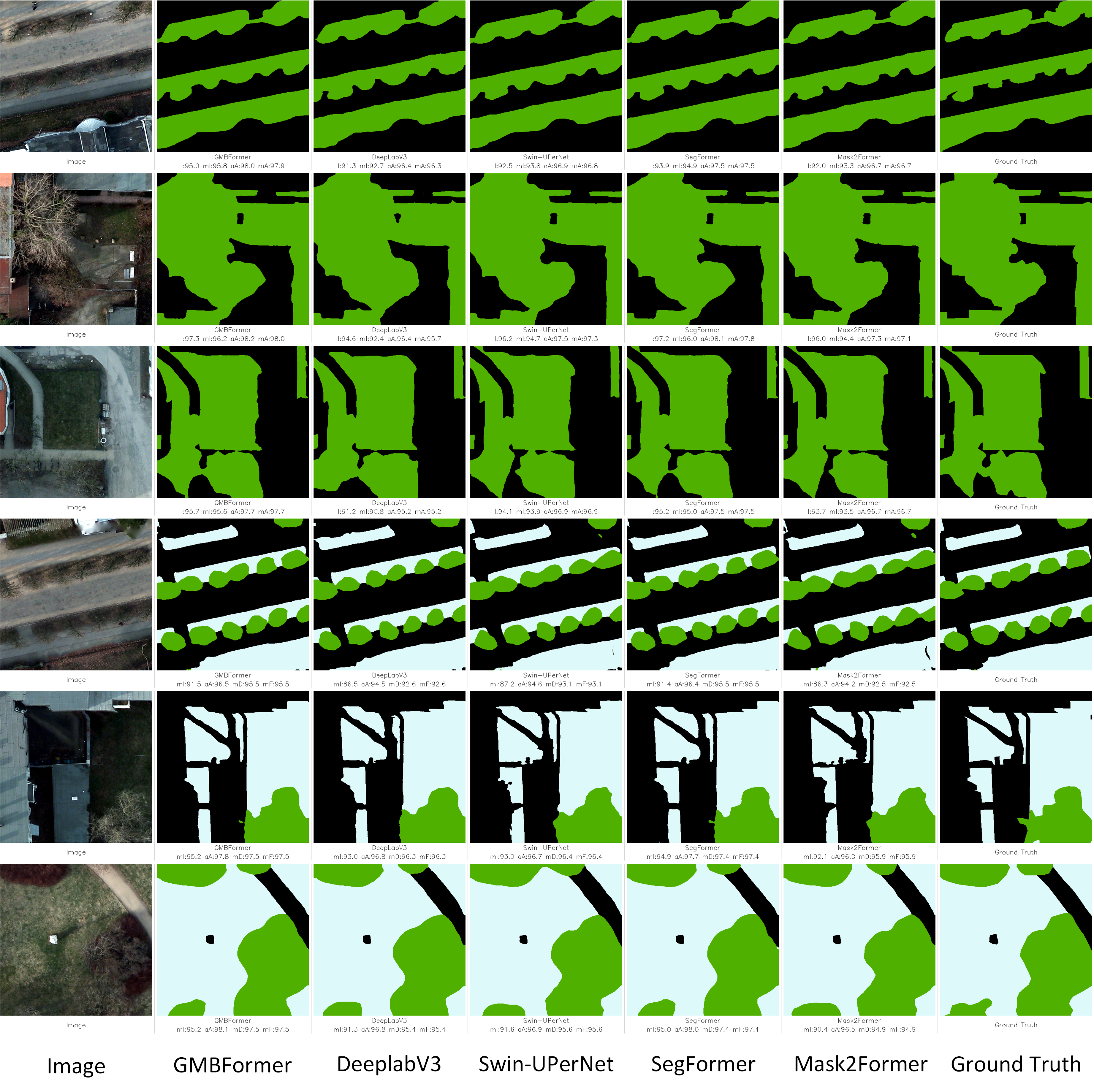}
  \caption{Qualitative comparison on ISPRS Potsdam binary and three-class green-space extraction results.}
  \label{fig:potsdam_qualitative_comparison}
\end{figure}

Fig.~\ref{fig:potsdam_qualitative_comparison} supports the quantitative results from both Potsdam settings. In binary maps, GMBFormer follows elongated vegetation strips and small green patches more continuously. In three-class maps, it better preserves the distinction between low vegetation and tree crowns, visually indicating that retrieved prototypes strengthen vegetation structure without collapsing the two subclasses into a single foreground.

\subsection{Qualitative Map-Level Analysis}

\begin{figure}[H]
  \centering
  \includegraphics[width=0.95\linewidth]{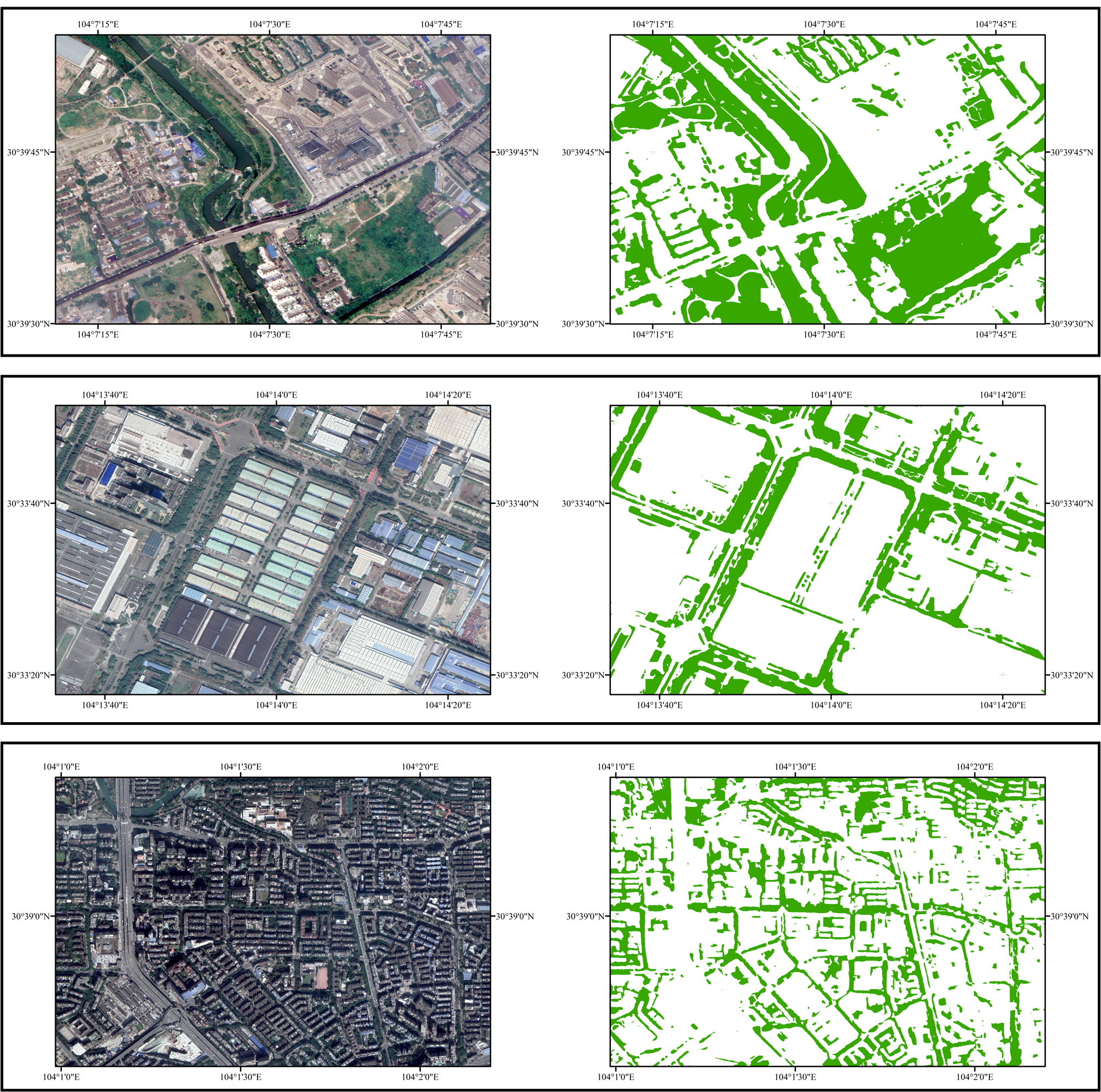}
  \caption{Map-level green-space extraction in three built-up regions of Chengdu.}
  \label{fig:qualitative_map}
\end{figure}

Fig.~\ref{fig:qualitative_map} evaluates map-level usability under practical urban mapping conditions. The examples cover park-dominated vegetation, roadside linear greenery, and fragmented greenery within dense buildings.

The predictions preserve contiguous parks, narrow roadside vegetation, and scattered courtyard greenery, suggesting that memory-guided retrieval is helpful for multiple urban vegetation morphologies rather than a single scene type.

\subsection{Ablation Study}

Table~\ref{tab:ablation} summarizes the ablation study on the Chengdu validation set. The RGBA variant denotes red--green--blue--alpha input, with NDVI stored in the alpha channel.

\begin{table}[H]
\centering
\caption{Ablation study results on the Chengdu validation set.}
\label{tab:ablation}
\scriptsize
\setlength{\tabcolsep}{4pt}
\resizebox{\textwidth}{!}{%
\begin{tabular}{lcccccccc}
\toprule
Variant & $S$ & $\tau_{\text{raw}}$ & $\alpha$ & NDVI Gate & Memory Bank & mIoU & mDice & IoU (green) \\
\midrule
M1 -- SegFormer-B4 (RGB only) & -- & -- & -- & \xmark & \xmark & 87.40 & 92.83 & 84.75 \\
M2 -- RGBA concat & -- & -- & -- & \xmark & \xmark & 86.52 & 91.97 & 84.82 \\
M3 -- No gate & 64 & -- & 0.99 & \xmark & \cmark & 87.37 & 92.82 & 85.64 \\
M4 Full (Ours) & 64 & 0.2 & 0.99 & \cmark & \cmark & 89.25 & 94.31 & 86.76 \\
$\tau_{\text{raw}} = 0.1$ & 64 & 0.1 & 0.99 & \cmark & \cmark & 87.61 & 93.01 & 85.96 \\
$\tau_{\text{raw}} = 0.3$ & 64 & 0.3 & 0.99 & \cmark & \cmark & 87.70 & 93.10 & 86.05 \\
$\tau_{\text{raw}} = 0.5$ & 64 & 0.5 & 0.99 & \cmark & \cmark & 87.53 & 92.94 & 85.88 \\
$S = 32$ & 32 & 0.2 & 0.99 & \cmark & \cmark & 87.65 & 93.05 & 86.00 \\
$S = 128$ & 128 & 0.2 & 0.99 & \cmark & \cmark & 87.74 & 93.14 & 86.09 \\
$\alpha = 0.9$ & 64 & 0.2 & 0.9 & \cmark & \cmark & 88.06 & 93.64 & 86.28 \\
$\alpha = 0.999$ & 64 & 0.2 & 0.999 & \cmark & \cmark & 87.58 & 92.99 & 85.93 \\
\bottomrule
\end{tabular}
}
\end{table}

Direct RGBA concatenation (M2) degrades mIoU relative to the RGB-only baseline (M1), indicating that simply treating NDVI as a fourth learnable channel is not an effective use of the physical cue in this setting. The ungated memory variant (M3) improves green-space IoU over M1 but does not improve mIoU, suggesting that memory retrieval needs admission control to avoid unbalanced prototype updates. The full model combines NDVI-guided admission and memory retrieval, raising mIoU to 89.25\% and green-space IoU to 86.76\%. The parameter ablations further show that $\tau_{\text{raw}}=0.2$, $S=64$, and $\alpha=0.99$ provide a favorable operating point: looser or stricter admission, smaller or larger memory capacity, and faster or slower EMA updates all reduce the benefit.

\subsection{Computational Cost}

\begin{table}[H]
\centering
\caption{Computational cost comparison for a single $512 \times 512$ patch. FLOPs denotes floating-point operations.}
\label{tab:cost}
\scriptsize
\setlength{\tabcolsep}{5pt}
\resizebox{\textwidth}{!}{%
\begin{tabular}{llccc}
\toprule
Model & Backbone & Parameters (M) & FLOPs (G) & Inference (ms/patch) \\
\midrule
DeepLabV3 & ResNet-50-D8 & 65.74 & 269.81 & 70.10 \\
SegFormer-B4 & MiT-B4 & 61.37 & 58.57 & 28.35 \\
Swin-UPerNet & Swin-Base & 120.12 & 304.69 & 73.64 \\
Mask2Former & ResNet-50 & 44.00 & 66.32 & 32.78 \\
GMBFormer (Ours) & MiT-B4 & 61.88 & 58.94 & 28.48 \\
\bottomrule
\end{tabular}
}
\end{table}

As shown in Table~\ref{tab:cost}, GMBFormer introduces only a small structural overhead over the SegFormer-B4 backbone, increasing parameters by 0.51~M and FLOPs by 0.37~G. The measured single-patch latency remains close to SegFormer-B4 (28.48 versus 28.35~ms), but this sub-millisecond difference should be interpreted as comparable runtime under the present setup rather than as a precise latency estimate. Its computational cost remains close to SegFormer-B4 and is substantially lower than DeepLabV3 and Swin-UPerNet in FLOPs, while the preceding comparisons show favorable segmentation accuracy. Mask2Former has fewer parameters, but its measured latency is higher than GMBFormer under the same single-patch evaluation setting.

\section{Discussion}
\label{sec:discussion}

\subsection{Why Decoupling NDVI from the Backbone Works}

The comparison between red--green--blue--alpha (RGBA) concatenation (M2) and GMBFormer (M4 Full) suggests that, for this task, NDVI is more useful as a physical prior for memory control than as an additional learnable input channel. Direct concatenation perturbs the ImageNet-pretrained RGB stem and forces the backbone to jointly model appearance and vegetation-index distributions. GMBFormer instead keeps RGB responsible for representation learning and uses NDVI only to control memory admission, preserving the pretrained RGB prior while retaining the ecological meaning of NDVI as vegetation confidence.

\subsection{Effect of NDVI Resolution Mismatch}

For the Chengdu dataset, NDVI is derived from 10~m Sentinel-2 observations and then aligned to 0.27~m Google Earth imagery. This cross-resolution construction cannot preserve object-level vegetation boundaries in the same way as the UHR RGB image, especially around narrow roadside greenery, small courtyard vegetation, and mixed pixels near building edges. In GMBFormer, however, this mismatch affects only the memory write gate: NDVI determines whether a patch-level RGB prototype is admitted to the memory bank, while all feature extraction, retrieval queries, decoding, and pixel-level prediction remain RGB-based. Therefore, NDVI misalignment may introduce admission noise, such as rejecting small vegetation-dominated patches or admitting mixed patches, but it does not directly inject coarse Sentinel-2 texture into the learned feature maps. Using patch-mean NDVI further makes the gate less sensitive to local boundary misregistration, at the cost of losing fine within-patch vegetation structure. The threshold and capacity ablations partly reflect this trade-off, and higher-resolution NIR observations or adaptive confidence estimation could further reduce gate noise.

\subsection{The Memory Bank as a Semantic Bridge across Patches}

The memory bank externalizes training-set vegetation experience into a similarity-indexed prototype dictionary. The slot assignment in Eq.~\eqref{eq:prototype_matching} and EMA update in Eq.~\eqref{eq:ema_update} act like online spherical clustering, encouraging slots to represent recurring vegetation appearances without explicit clustering supervision. During reading, a patch retrieves prototypes by semantic similarity rather than spatial adjacency, which is important in urban scenes where neighboring crops can contain unrelated land-cover types and visually similar greenery may appear far apart.

\subsection{Generalizability on Potsdam-Derived Settings}

The Chengdu validation and Potsdam-derived results indicate that the proposed mechanism is not limited to the Chengdu imagery. The NDVI gate reduces the chance that RGB-specific distractors enter the memory bank, while the stored prototypes are intended to capture higher-level vegetation structure rather than only local color statistics. The same write-gate-read design could be explored for other UHR segmentation tasks when a reliable physical index is available, such as the Normalized Difference Built-up Index (NDBI) for buildings or the Normalized Difference Water Index (NDWI) for water bodies.

\subsection{Limitations and Future Work}

Several limitations remain. First, the memory bank reflects the training distribution, so rare vegetation appearances may be underrepresented. Second, the capacity $S$ is fixed empirically rather than adjusted by prototype utilization; future work should quantify slot usage and adapt capacity automatically. Third, the current memory is attached only to Stage--3 features. Fourth, coarse-resolution NDVI can introduce uncertainty into memory admission even though it is excluded from feature learning. Finally, NDVI requires NIR information, which may be unavailable for RGB-only imagery; learning a reliable RGB-based vegetation-confidence substitute remains future work.

\section{Conclusion}
\label{sec:conclusion}

This study introduced GMBFormer for urban green-space extraction from patch-based UHR imagery. The method decouples RGB representation learning from NDVI-guided memory admission, stores high-confidence vegetation prototypes in a global memory bank, and retrieves them through cross-attention to enhance current patch features by semantic similarity rather than spatial adjacency.

Experiments on the Chengdu UHR dataset and two reduced-label settings derived from ISPRS Potsdam show improvements over representative segmentation baselines retrained under the same protocols. Ablations indicate that direct RGBA concatenation is insufficient, ungated memory is less reliable, and the final performance depends on an appropriate admission threshold, memory size, and EMA momentum. Future work will explore adaptive memory capacity, multi-scale memory, and vegetation-confidence estimation when NIR-derived NDVI is unavailable.

\section*{Declaration of Competing Interest}
The authors declare that they have no known competing financial interests or personal relationships that could have appeared to influence the work reported in this paper.

\section*{Data and Code Availability}
The data and source code necessary to reproduce the results of this study are publicly available at \url{https://github.com/xicheng79/GMBFormer}. The ISPRS Potsdam dataset used for additional evaluation is publicly available from the ISPRS 2D Semantic Labeling Challenge.

\section*{Acknowledgements}
This research was supported by the Chengdu University of Technology ``AI Research Fund'' Program (2025AI038), the Yunnan Sci-Tech Talent and Platform Plan Project (202605AF350036), and the ``Engineering Research Center for Innovation Application of Low-Altitude Remote Sensing in Plateau and Mountainous Regions (Incubation)'' Project of the Yunnan Province Department of Education.

\end{document}